%% file: main.tex
\documentclass[sigconf,natbib=true]{acmart}
\AtBeginDocument{%
  }

\setcopyright{acmlicensed}
\copyrightyear{2026}
\acmYear{2026}
\acmDOI{XXXXXXX.XXXXXXX}

\acmConference[SIGIR '26]{The 49th International ACM SIGIR Conference on Research and Development in Information Retrieval}{July 20–24, 2026}{Melbourne | Naarm, Australia}

\acmISBN{978-1-4503-XXXX-X/2018/06}

\usepackage{colortbl}
\usepackage{listings}
\usepackage{booktabs}
\usepackage{caption}
\usepackage{multirow}
\usepackage{enumitem}
\usepackage{algorithm}
\usepackage{algpseudocode}
\usepackage{amsmath}
\usepackage[table]{xcolor}

\begin{document}

\title{STRIDE: Strategic Iterative Decision-Making for Retrieval-Augmented Multi-Hop Question Answering}

\author{Wei Chen}
\orcid{0000-0003-4700-2612}
\affiliation{%
  \institution{University of Science and Technology of China \& State Key Laboratory of Cognitive Intelligence}
  \city{Hefei}
  \state{Anhui}
  \country{China}
}
\email{chenweicw@mail.ustc.edu.cn}

\author{Lili Zhao}
\orcid{0000-0002-5786-2424}
\affiliation{%
  \institution{University of Science and Technology of China \& State Key Laboratory of Cognitive Intelligence}
  \city{Hefei}
  \state{Anhui}
  \country{China}
}
\email{liliz@mail.ustc.edu.cn}

\author{Zhi Zheng}
\authornote{Corresponding authors}
\orcid{0000-0001-7758-8904}
\affiliation{%
  \institution{University of Science and Technology of China \& State Key Laboratory of Cognitive Intelligence}
  \city{Hefei}
  \state{Anhui}
  \country{China}
}
\email{zhengzhi97@ustc.edu.cn}

\author{Huijun Hou}
\affiliation{%
 \institution{NIO}
 \city{Hefei}
  \state{Anhui}
  \country{China}
}
\email{huijun.hou@nio.com}

\author{Tong Xu}
\authornotemark[1]
\orcid{0000-0003-4246-5386}
\affiliation{%
  \institution{University of Science and Technology of China \& State Key Laboratory of Cognitive Intelligence}
  \city{Hefei}
  \state{Anhui}
  \country{China}
}
\email{tongxu@ustc.edu.cn}

\renewcommand{\shortauthors}{Trovato et al.}

\begin{abstract}
   Multi-hop question answering (MHQA) enables accurate answers to complex queries by retrieving and reasoning over evidence dispersed across multiple documents.
   Existing MHQA approaches mainly rely on iterative retrieval-augmented generation, which suffer from the following two major issues. On one hand, existing methods prematurely commit to surface-level entities rather than underlying reasoning structures, making question decomposition highly vulnerable to lexical ambiguity. On the other hand, existing methods overlook the logical dependencies among reasoning steps, resulting in uncoordinated execution.
   To address these issues, we propose \textbf{STRIDE} (\textbf{Str}ategic \textbf{I}terative \textbf{De}cision-making), a framework that separates strategic planning, dynamic control, and grounded execution.
   At its core, a \textit{Meta-Planner} first constructs an entity-agnostic reasoning skeleton to capture the abstract logic of the query, thereby deferring entity grounding until after the reasoning structure is established, which mitigates disambiguation errors caused by premature lexical commitment.
   A \textit{Supervisor} then orchestrates sub-question execution in a dependency-aware manner, enabling efficient parallelization where possible and sequential coordination when necessary. By dynamically deciding whether to retrieve new evidence or infer from existing facts, it avoids redundant queries and error propagation, while fusing cross-branch information and reformulating failed queries to enhance robustness.  
   Grounded fact extraction and logical inference are delegated to specialized execution modules, ensuring faithfulness through explicit separation of retrieval and reasoning.
   While STRIDE is compatible with any large language models (LLMs), off-the-shelf open-source LLMs underperform closed-source counterparts in its structured reasoning pipeline. To close this gap, we further propose \textbf{STRIDE-FT}, a modular fine-tuning framework that uses self-generated execution trajectories from STRIDE, requiring neither human annotations nor stronger teacher models. Experiments show that STRIDE achieves robust and accurate reasoning on MHQA benchmarks, while STRIDE-FT effectively enhances open-source LLMs\footnote{The code repository is available at \href{https://github.com/fanshu6hao/STRIDE}{github.com/fanshu6hao/STRIDE}.}.
\end{abstract}





\maketitle

\input{sections/01_intro}

\input{sections/02_related_work}

\input{sections/03_method}

\input{sections/04_experiment}

\input{sections/05_conclusion}

\section{Acknowledgments}
This work was supported in part by the grants from National Science and Technology Major Project (No. 2023ZD0121104), in part by the National Natural Science Foundation of China (No.62222213, U22B2059), in part by the Postdoctoral Fellowship Program and China Postdoctoral Science Foundation under Grant Number BX20250387 and 2025M781529, in part by the Fundamental Research Funds for the Central Universities under Grant Number WK2150250042. And this work was also supported by USTC-NIO Smart Electric Vehicle Joint Lab.

\bibliographystyle{ACM-Reference-Format}
\bibliography{citations}

\end{document}

%% file: sections/01_intro.tex
\begin{figure}[htbp]
\centering
\includegraphics[width=.9\linewidth]{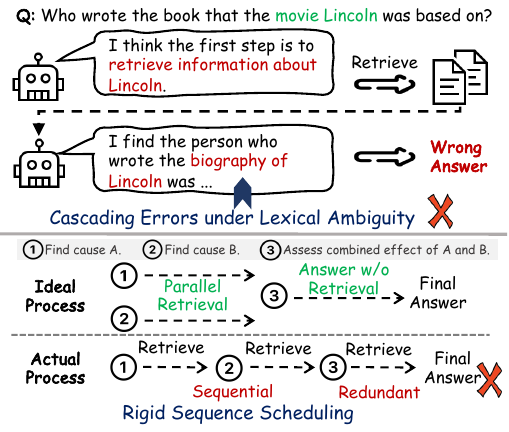}
\vspace{-10pt}
\caption{
Challenges in current iterative RAG. Top: Premature entity grounding leading to cascading errors due to lexical ambiguity. Bottom: Rigid scheduling that fails to account for the rich dependency structures among sub-questions.
}
\label{fig:intro}
\vspace{-10pt}
\end{figure}

\section{Introduction}
Open-domain question answering (QA) aims to answer user questions using a large unstructured corpus without being restricted to a predefined knowledge base.
Within this setting, multi-hop question answering (MHQA) represents a core challenge, requiring systems to synthesize information scattered across multiple documents through coherent and multi-step reasoning~\cite{yang2024large, zhong2023mquake, xu2025harnessing, chen2025following}.
Unlike simple factual queries resolvable via single-document lookup, real-world user questions such as “What Nobel Prize-winning scientist was also an accomplished concert musician?” demand not only accurate evidence retrieval but also structured orchestration of interdependent reasoning steps~\cite{zhang-etal-2024-end,lan2021survey,sun2024harnessing}.

Retrieval-Augmented Generation (RAG) has become a dominant paradigm for open-domain QA by grounding large language model (LLM) generation in external knowledge, thereby improving factuality and extending coverage beyond parametric memory~\cite{xu2026from,cuconasu2024power,xu2024list,edge2024local,DBLP:conf/sigir/0007RWZ00025}.
Standard RAG relies on a \textit{single} retrieval step using the original query. While this works well for simple factual questions, it often falls short on multi-hop questions, where reasoning steps are interdependent and the initial query retrieves only sparsely relevant documents~\cite{fan2024survey}.
To bridge this gap, recent works have turned to \textit{iterative RAG} frameworks~\cite{selfrag, dualrag, trivedi2023interleaving, metarag, activerag, planrag}, which decompose the question into sub-queries and alternate between retrieval and reasoning over multiple turns.  
For example, to answer “Who directed the film based on George Orwell’s novel 1984?”, an iterative RAG system might first retrieve information about the novel’s adaptations, then use that context to formulate a follow-up query about the director of the specific film version.

Despite their progress, existing iterative RAG approaches suffer from two fundamental limitations as shown in Figure~\ref{fig:intro}.
First, they often perform entity-centric decomposition too early, before establishing a task-aligned reasoning structure. This can cause cascading errors when entity names are ambiguous.
For example, in the question ``Who wrote the book that the movie Lincoln was based on?'', the model may prematurely ground on the surface entity ``Lincoln'' and retrieve information about Abraham Lincoln rather than the film ``Lincoln''. 
As a result, it might answer with the author of a biography of Abraham Lincoln instead of the screenwriter of the movie’s source book.
In contrast, if the system first abstracts the reasoning path as [movie → source book → author], it can correctly disambiguate ``Lincoln'' as a film entity and avoid being misled by homonymous references.

Second, existing approaches treat multi-hop reasoning as a rigid, sequential pipeline, ignoring the rich dependency structures inherent in complex questions. In reality, sub-questions can exhibit diverse relationships: some are strictly sequential (e.g., ``Find entity A'' → ``Find property of A''), others are independent and parallelizable (e.g., ``Find birth year of X'' and ``Find birth year of Y'' for comparison), and yet others form fork-join patterns (e.g., ``Find causes P and Q'' → ``Assess combined effect of P and Q''). 
Without explicit modeling of such dependencies, current systems either waste resources through unnecessary sequential execution, issue redundant or conflicting queries, or fail to integrate evidence across reasoning branches.
Moreover, they lack the ability to dynamically determine based on already gathered facts whether a sub-question necessitates external retrieval or can be resolved through logical inference, leading to inefficient resource usage, error propagation, and brittle behavior when partial evidence is missing.

To address these limitations, we draw inspiration from real-world decision-making hierarchies: \textit{strategic leaders define global goals, middle managers assess progress and schedule tasks, and operational agents execute concrete actions}. Mirroring this tripartite structure, we propose \textbf{STRIDE} (\textbf{Str}ategic \textbf{I}terative \textbf{De}cision-making), a framework that explicitly separates reasoning into strategy, control, and execution.
At the \textbf{\textit{strategy layer}}, the Meta-Planner enforces a reasoning-first perspective by decoupling logical structure from linguistic surface form. Rather than decomposing the question directly into entity-bound sub-queries, it first constructs a \textit{General Strategy}, which is an abstract representation of the reasoning flow that is independent of specific entity names. Only after this high-level logic is established does it instantiate a \textit{Concrete Plan} of executable, entity-grounded sub-questions. This two-stage design prioritizes \textit{how to reason} before committing to \textit{what to retrieve}, thereby reducing sensitivity to lexical ambiguity in entity references.
At the \textbf{\textit{control layer}}, the Supervisor acts as process-aware orchestrator that schedules sub-questions according to their logical dependencies. 
Guided by the \textit{Concrete Plan}, it dynamically selects execution modes (e.g., sequential, parallel, or fork-join) and decides at each step whether to invoke retrieval or apply inference over existing evidence.
Throughout execution, it monitors retrieval quality, fuses cross-branch information, refrains from redundant queries when answers are inferable, and reformulates semantically ambiguous requests to recover from failed retrievals.
At the \textbf{\textit{execution layer}}, dedicated modules carry out the Supervisor’s instructions. The Extractor collects atomic facts from external sources, while the Reasoner performs logical inference over gathered facts. 
Their tight integration ensures both interpretability and output faithfulness.

While STRIDE is model-agnostic and compatible with both open and closed LLMs, practical deployment often favors open-source models due to cost and privacy constraints. However, these models generally underperform their closed-source counterparts in the structured multi-step reasoning required by STRIDE. To bridge this gap, we further introduce \textbf{STRIDE-FT}, a modular fine-tuning framework that leverages self-executed trajectories from STRIDE to train each component, enabling effective adaptation of open-source LLMs without human labels or stronger teacher models.

In summary, our contributions are threefold:
\begin{itemize}[leftmargin=*]
    \item We propose \textbf{STRIDE}, a decision-inspired framework for multi-hop QA that explicitly separates \textit{strategy}, \textit{control}, and \textit{execution}. It enables robust improvement through reasoning-first planning and dynamic dependency-aware scheduling.
    
    \item We introduce the \textbf{Meta-Planner} and \textbf{Supervisor} as core modules. The Meta-Planner constructs an entity-agnostic reasoning skeleton before grounding, mitigating ambiguity-induced errors. The Supervisor dynamically schedules dependency-aware sub-question execution and integrates mechanisms for query reformulation, redundancy avoidance, and cross-branch evidence fusion.
    
    \item We present \textbf{STRIDE-FT}, a modular fine-tuning framework that leverages self-executed trajectories from STRIDE to train each component. STRIDE-FT enables effective adaptation of open-source LLMs without human labels or stronger teacher models.
\end{itemize}

%% file: sections/02_related_work.tex
\section{Related Work}
\label{related_work}

\noindent\textbf{Multi-Hop Question Answering.}
Current models achieved strong performance on single-hop questions, yet struggled when answers required synthesizing information across multiple documents or reasoning steps~\cite{yang2018hotpotqa, wei2022chain, khalifa2023few}.
Early approaches addressed this by constructing entity- or evidence-centric graphs and applying graph neural networks to model inter-document relationships~\cite{ding2019cognitive, fang2020hierarchical}.
More recently, large language models (LLMs) have been leveraged for their strong textual reasoning capabilities~\cite{zhao2025evaluating,li2026dynadebate,lyu2026mock}, often augmented with explicit information extraction or evidence composition modules to capture cross-entity dependencies~\cite{structqa, holmes,chen2024double,xu2024large}.
However, most of these methods assumed access to \textit{gold} supporting documents, sidestepping the realistic challenge of retrieving relevant evidence from a large open corpus. In contrast, our work operates in a fully open-domain setting, where both retrieval and multi-step reasoning must be jointly orchestrated—a more practical and challenging scenario.

\noindent\textbf{Retrieval-Augmented Generation (RAG).}
Despite their impressive capabilities, LLMs are prone to hallucination on factual queries due to the static and bounded nature of their parametric knowledge~\cite{zhao2023survey, DBLP:conf/sigir/YangRCGZYZ24,wang2025think}.
To mitigate this, Retrieval-Augmented Generation (RAG) has emerged as a dominant paradigm, grounding model responses in external knowledge retrieved at inference time~\cite{lewis2020retrieval, izacard2023atlas, DBLP:conf/sigir/SalemiZ24a, DBLP:conf/sigir/WangTCS25}.
Standard RAG performed a \textit{single} retrieval step using original query, followed by answer generation~\cite{gao2023retrieval, cuconasu2024power}.
While effective for simple factoid questions, this approach often failed on multi-hop queries, as the initial retrieval tended to miss critical intermediate evidence, leading to incomplete or incorrect reasoning~\cite{fan2024survey}.

\noindent\textbf{Iterative RAG.}
To overcome the limitations of single-step retrieval, recent works adopted \textit{iterative RAG} frameworks that alternate between query decomposition, retrieval, and reasoning over multiple turns~\cite{trivedi2023interleaving, metarag, activerag}.
For instance, PlanRAG~\cite{planrag} first generated a reasoning plan and then executed it step-by-step, while SelfAsk~\cite{press2023measuring} decomposed questions by sequentially asking follow-up questions and using retrieved information to refine answers iteratively.
\begin{figure*}[!t]
  \centering
  \includegraphics[width=0.9\linewidth]{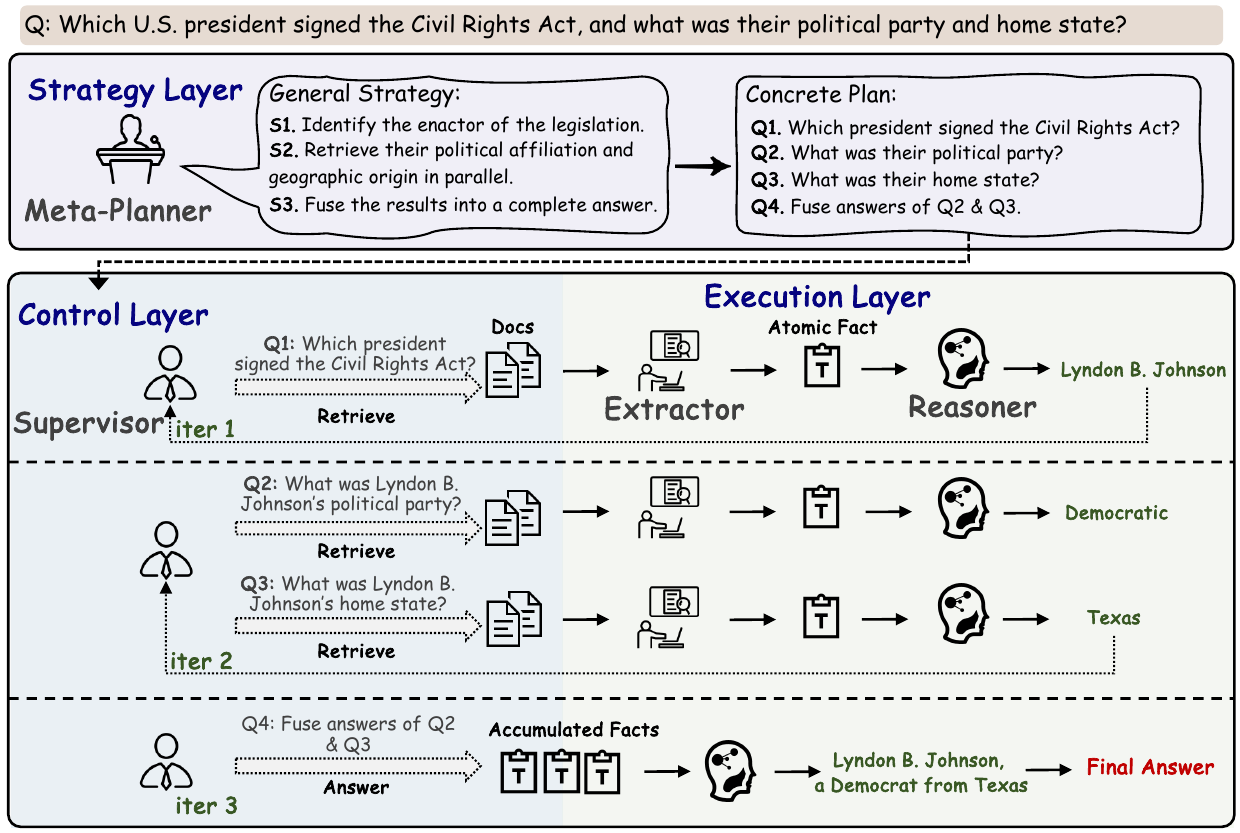}
  \caption{Overview of STRIDE, which structures the reasoning process into strategy, control, and execution.}
  \vspace{-10pt}
  \label{fig:framework}
\end{figure*}
Iter-RetGen~\cite{shao2023enhancing} leveraged iterative cycles where previous generations are concatenated with the query to retrieve more relevant information, thus refining subsequent inferences progressively.
GenGround~\cite{genground} iteratively generated sub-questions and answers, and then retrieved documents to revise the answers, leveraging the model’s generative capability to guide retrieval.
DualRAG~\cite{dualrag} maintained a dual process to integrate reasoning and retrieval for contiguous reasoning.
Despite the progress, they typically decomposed questions directly into entity-bound sub-queries and executed steps in a fixed sequential order, overlooking abstract reasoning structure and rich dependency patterns. 
In contrast, STRIDE decouples reasoning into strategy, control, and execution, enabling abstract planning, dynamic scheduling, and robust recovery in MHQA.

%% file: sections/03_method.tex
\section{Methodology}
In this section, we present \textbf{STRIDE} (\textbf{Str}ategic \textbf{I}terative \textbf{De}cision-making), a framework that models MHQA as a structured decision-making process. 
As illustrated in Figure~\ref{fig:framework}, STRIDE structures the reasoning process into three tightly coordinated layers (i.e., strategy, control, and execution), enabling global planning, adaptive scheduling, and grounded inference. 
We detail the architecture of STRIDE in Section~\ref{method:framework}, and describe how its self-executed trajectories can be leveraged for modular fine-tuning in Section~\ref{sec:modular_tuning}.

\subsection{Task Definition}
We formally define the Retrieval-Augmented Generation (RAG) task as follows: 
Given a question $q$ and a document corpus $D=\{d_i\}_{i=1}^{|D|}$, the task demands to derive a prediction $y$ by retrieving relevant documents $D_q \subseteq D$ and capturing the useful information.
It can be formally defined as:
\begin{equation}\label{eq1}
 y = M_{\theta}(I, q, D_q), 
\end{equation}
where $M_{\theta}$ denotes the LLM model, and $I$ is the task instruction.

\subsection{Framework of STRIDE}
\label{method:framework}
Inspired by real-world decision hierarchies, STRIDE models multi-hop QA as a structured decision-making process composed of three layers. 
The \textit{strategy layer} first constructs an abstract, entity-agnostic reasoning skeleton before grounding it into concrete sub-questions, ensuring global coherence and reducing sensitivity to lexical ambiguity. 
The \textit{control layer} acts as an adaptive supervisor that dynamically schedules sub-questions based on logical dependencies, supports sequential, parallel, and fork-join execution patterns, and decides per step whether to retrieve new evidence or reason over existing facts. 
Finally, the \textit{execution layer} performs grounded fact extraction and faithful logical inference using dedicated modules. 
This process departs from flat, uncoordinated RAG pipelines, enabling reasoning that is both deliberate and recoverable.

\subsubsection{Strategy Layer}
To address the challenge of compositional reasoning in multi-hop question answering, we introduce \textbf{\textit{Meta-Planner}}, a reasoning planner that decomposes a question into a structured, two-level reasoning blueprint. This design explicitly separates abstract reasoning logic from entity-specific instantiation, thereby reducing sensitivity to surface-form variations of complex entities (e.g., film titles, administrative names) and enhancing plan transferability across questions with shared reasoning patterns.

Formally, given a question \( q \), the Meta-Planner \( M_{\text{plan}} \) generates a structured reasoning blueprint \( P_q \) consisting of two levels:
\begin{equation}
P_q = \{ S_q, C_q \} = M_{\text{plan}}(I_{\text{plan}}, q),   
\end{equation}
where \( I_{\text{plan}} \) denotes the planning instruction, 
\( S_q = \{ s_1, s_2, \ldots, s_m \} \) is the \textit{General Strategy}—an abstract sequence of reasoning steps expressed over entity \textit{types} (e.g., \textit{person}, \textit{location}, \textit{work}), independent of specific names. And
\( C_q = \{ c_1, c_2, \ldots, c_n \} \) is the \textit{Concrete Plan}—a minimal, ordered set of executable sub-questions grounded in the entities and relations from \( q \), derived by instantiating \( S_q \).

Hence, the Meta-Planner serves as the strategic layer of STRIDE, producing interpretable, logically structured reasoning plans that guide subsequent layers in a top-down fashion.

\subsubsection{Control Layer}
\label{sec:supervisor}

While the Meta-Planner produces an ordered concrete plan $C_q$, actual execution must handle: (1) sub-questions may depend on prior answers and cannot be executed out-of-order; (2) retrieval for fact-seeking steps can fail due to lexical mismatch or sparse evidence, requiring adaptive reformulation; and (3) some steps (e.g., set intersection or comparison) require no retrieval at all, but pure reasoning over known facts. 
To address these challenges, the \textbf{\textit{Supervisor}} serves as a process-aware controller that dynamically orchestrates sub-question execution. By maintaining a global view of progress, inter-step dependencies, and past retrieval outcomes, it makes context-sensitive decisions on scheduling, retrieval invocation, and reasoning strategy.

Specifically, the Supervisor maintains an execution state denoted by 
$\Omega = (\mathcal{S}, \mathcal{P}, \mathcal{F})$, where 
$\mathcal{S} \subseteq C_q$ is the set of solved sub-questions with their answers, 
$\mathcal{P} \subseteq C_q$ is the set of pending sub-questions (with $\mathcal{S} \cap \mathcal{P} = \emptyset$), 
and $\mathcal{F}$ is a partial map that, for each pending sub-question $c_i \in \mathcal{P}$, records the set of retrieval queries that yielded no useful results.
At each step, it identifies the set of ready sub-questions $\mathcal{P}_{\text{ready}} \subseteq \mathcal{P}$ whose dependencies (i.e., required answers in $\mathcal{S}$) are satisfied, and invokes an instruction-following LLM to determine the next actions:
\begin{equation}
    \mathcal{D} = M_{\text{super}}(I_{\text{super}}, q, C_q, \Omega),
\end{equation}
where $I_{\text{super}}$ is the task instruction, and $\mathcal{D} = \{( \alpha_i, z_i) \mid c_i \in \mathcal{P}_{\text{ready}}\}$ specifies the action $\alpha_i \in \{\texttt{retrieve}, \texttt{rewrite}, \texttt{answer}\}$ and associated query $z_i$ for each ready sub-question $c_i$.

The decision is guided by three scheduling principles:
\begin{enumerate}[leftmargin=*,nosep]
    \item If $c_i$ is fact-seeking and $\mathcal{F}(c_i) = \emptyset$, issue \texttt{retrieve} with a concise, entity-grounded query $z_i$;
    \item If $\mathcal{F}(c_i) \neq \emptyset$, generate a \texttt{rewrite} via paraphrasing or semantic reformulation, ensuring $z_i \notin \mathcal{F}(c_i)$;
    \item If $c_i$ requires only reasoning over known answers (e.g., intersection, comparison), trigger \texttt{answer}—bypassing retrieval and delegating to the reasoning module.
\end{enumerate}

Critically, the Supervisor enables \textit{parallel execution} of independent sub-questions (e.g., $c_1$ and $c_2$ in a set-intersection question), while deferring dependent steps (e.g., $c_3$) until their prerequisites are resolved. 
The Supervisor operates within an iterative RAG loop: at each iteration $t$, it emits $\mathcal{D}$, which is consumed by the execution layer to resolve each $c_i \in \mathcal{D}$; the resulting answers and failure signals update $\Omega$, and the loop continues until $\mathcal{P} = \emptyset$ or a maximum iteration limit $T_{\max}$ is reached.

\subsubsection{Execution Layer}
\label{sec:execution}

The Execution Layer realizes the Supervisor’s scheduling decisions $\mathcal{D}$ by invoking two specialized modules. An \textbf{\textit{Extractor}} for grounding atomic facts in retrieved documents and a \textbf{\textit{Reasoner}} for synthesizing final answers from facts. 
For each directive $(\alpha_i, z_i) \in \mathcal{D}$:
\begin{itemize}[leftmargin=*,nosep]
    \item if $\alpha_i \in \{\texttt{retrieve}, \texttt{rewrite}\}$, the system first retrieves top-$k$ documents $D_{z_i} \subseteq D$ using a dense retriever, invokes the Extractor to distill atomic facts $\mathcal{E}_i$ from $D_{z_i}$, and then passes $\mathcal{E}_i$ to the Reasoner to answer $z_i$;
    \item if $\alpha_i = \texttt{answer}$, the Reasoner directly answers $z_i$ using all facts accumulated from previous extractions.
\end{itemize}
The outputs are used to update the execution state $\Omega$: successful answers are added to $\mathcal{S}$, while failed retrievals (e.g., empty fact sets) trigger updates to $\mathcal{F}$.

Both modules are implemented as instruction-following LLMs:
\begin{align}
    \text{Extractor:} \quad & \mathcal{E}_i = M_{\text{ext}}(I_{\text{ext}}, z_i, D_{z_i}), \\
    \text{Reasoner:} \quad & a_i = M_{\text{rea}}(I_{\text{rea}}, z_i, \mathcal{X}_i),
\end{align}
where $\mathcal{X}_i = \mathcal{E}_i$ if $\alpha_i \in \{\texttt{retrieve}, \texttt{rewrite}\}$, and $\mathcal{X}_i$ denotes all facts accumulated so far (from prior extractions) if $\alpha_i = \texttt{answer}$. 
The prompt $I_{\text{ext}}$ instructs faithful, atomic, and relevant fact extraction, while $I_{\text{rea}}$ guides step-by-step reasoning over the provided evidence. 
If $\mathcal{E}_i = \emptyset$ after retrieval, the attempt is logged as a failure: $\mathcal{F}(c_i) \gets \mathcal{F}(c_i) \cup \{z_i\}$, enabling the Supervisor to trigger a \texttt{rewrite} in subsequent iterations.

In summary, the Execution Layer serves as the grounded reasoning engine of the system: by decoupling evidence extraction from answer synthesis, it ensures that every sub-question is resolved either through verifiable facts or explicit logical inference, providing the Supervisor with reliable, interpretable feedback to guide adaptive execution.

Moreover, to handle cases where the iterative process terminates without fully resolving the question (e.g., due to persistent retrieval failures, unparseable LLM outputs, or exceeding the iteration limit), we introduce a \textbf{\textit{Fallback Reasoner}}. This module takes as input the plan $P_q$, all facts collected during execution, and documents retrieved via the facts of last solved sub-question, and generates a best-effort answer to the original question $q$. This ensures the system always produces a response, improving robustness in practice.

The full inference procedure of STRIDE is formalized in Algorithm~\ref{alg:stride}. This executable pipeline not only produces answers but also generates rich, structured execution traces, enabling self-supervised modular training as described next.

\input{tables/algorithm_inference}

\subsection{Modular Fine-Tuning from Self-Executed Trajectories}
\label{sec:modular_tuning}

Unlike previous methods which tend to require stronger models as external supervision \cite{dualrag, genground}, we introduce \textbf{STRIDE-FT}, a modular fine-tuning framework that constructs training signals directly from the system’s own execution trajectories. 
For a held-out set of questions, we execute the full pipeline and log detailed traces, including the initial plan, per-step decisions, extracted facts, intermediate answers, and the final response. 
Crucially, we leverage the \textit{final outcome} (e.g., answer correctness, F1 score, iteration count, and failure flags) as an implicit signal to filter and refine module-specific inputs and outputs. 
Below, we detail the data construction strategy tailored to each component’s role.

\noindent\textbf{Meta-Planner: Preference Learning over Diverse Plans.}
To train the Meta-Planner to generate execution-friendly plans, we adopt a preference-based approach. For each question $q$, we sample $N=8$ diverse plans $\{\pi^{(i)}\}_{i=1}^N$ from the current planner and execute each through the full STRIDE pipeline. Each resulting trajectory is scored based on: (1) correctness of the final answer, (2) F1 overlap with ground truth, (3) number of iterations, and (4) presence of failure states (e.g., repeated rewrites). We then construct preference pairs $(\pi^+, \pi^-)$ from the same $q$ such that $\pi^+$ yields a significantly better outcome than $\pi^-$ (e.g., correct vs. incorrect, or F1 difference $>0.3$) and the two plans are sufficiently distinct (F1 score $<0.8$). These pairs are used to fine-tune the Meta-Planner via Direct Preference Optimization (DPO)~\cite{rafailov2023direct}.

\noindent\textbf{Supervisor: Learning Effective Rewrites.}
We focus on improving the \texttt{rewrite} action, which is critical when retrieval fails. We first collect all trajectories where the Supervisor issued a \texttt{rewrite} and the final answer was correct. From these, we filter out non-causal rewrites: (i) cases where the rewritten query was never executed (e.g., due to plan reordering), and (ii) cases where the action was labeled as \texttt{rewrite} but corresponded to the first retrieval attempt (i.e., effectively a \texttt{retrieve}). The remaining instances, in which a rewrite directly enabled successful fact extraction and sub-question resolution, form our training set. By training on this curated set of successful behaviors, the Supervisor learns to reproduce effective rewrites in similar contexts through Supervised Fine-Tuning (SFT).

\noindent\textbf{Extractor: Distilling Minimal Supporting Facts.}
The Extractor should output only facts necessary to answer the sub-question. We retain extraction outputs $\mathcal{E}_i$ only from trajectories with correct final answers. Among these, we focus on cases where $|\mathcal{E}_i| > 1$, indicating potential redundancy. We then use a same but frozen auxiliary LLM to select from $\mathcal{E}_i$ the minimal subset $\mathcal{E}_i^*$ that directly supports the answer to $c_i$, via the prompt: ``\textit{Given query: $z_i$, and candidate facts: $\mathcal{E}_i$, select only the facts strictly required to answer the question.}'' If $|\mathcal{E}_i^*| < |\mathcal{E}_i|$, the resulting $(z_i, D_{z_i}) \rightarrow \mathcal{E}_i^*$ pairs are collected to fine-tune the Extractor. 

\noindent\textbf{Reasoner: Promoting Concise and Deterministic Answers.}
The Reasoner sometimes produces verbose or ambiguous responses (e.g., listing multiple guesses), which harms answer precision. We select samples where: (i) the answer covers the ground truth with F1 $>0.3$, (ii) it is not an exact match, and (iii) its length exceeds the ground-truth answer by more than 50\%. To prevent the model from over-correcting toward excessive brevity or learning spurious patterns, we balance this set with twice as many randomly sampled examples where the original output already exactly matched the ground truth. All samples are trained via SFT to map $(z_i, \mathcal{X}_i)$ to a concise, single-statement answer aligned with the reference format.

This modular tuning strategy requires no human annotation or stronger teacher models. By aligning each component’s training objective with its functional role in the execution loop and filtering out non-causal or low-quality outputs, it enables targeted self-improvement while maintaining system coherence.

%% file: tables/algorithm_inference.tex
\begin{algorithm}[t]
\caption{The STRIDE Pipeline}
\label{alg:stride}
\begin{algorithmic}[1]
\Require Original question $q$, max iterations $T_{\max}$
\Ensure Final answer $a_{\text{final}}$

\State \textbf{// Strategy Layer: Generate reasoning blueprint}
\State $P_q \gets M_{\text{plan}}(I_{\text{plan}}, q)$ \Comment{Output: $(S_q, C_q)$}
\State $\mathcal{S} \gets \emptyset$, $\mathcal{P} \gets C_q$, $\mathcal{F} \gets \emptyset$ \Comment{Initialize execution state $\Omega$}
\State $t \gets 0$

\While{$\mathcal{P} \neq \emptyset$ \textbf{and} $t < T_{\max}$}
    \State $t \gets t + 1$
    
    \State \textbf{// Control Layer: Schedule ready sub-questions}
    \State $\mathcal{P}_{\text{ready}} \gets \{ c_i \in \mathcal{P} \mid \text{deps}(c_i) \subseteq \mathcal{S} \}$
    
    \State $\mathcal{D} \gets M_{\text{super}}(I_{\text{super}}, q, C_q, (\mathcal{S}, \mathcal{P}, \mathcal{F}))$
    \Comment{Output: $\{(\alpha_i, z_i)\}_{c_i \in \mathcal{P}_{\text{ready}}}$}
    
    \State \textbf{// Execution Layer: Resolve each directive}
    \For{each $(\alpha_i, z_i) \in \mathcal{D}$}
        \If{$\alpha_i \in \{\texttt{retrieve}, \texttt{rewrite}\}$}
            \State Retrieve top-$k$ docs $D_{z_i}$
            \State $\mathcal{E}_i \gets M_{\text{ext}}(I_{\text{ext}}, z_i, D_{z_i})$
            \If{$\mathcal{E}_i = \emptyset$}
                \State $\mathcal{F}(c_i) \gets \mathcal{F}(c_i) \cup \{z_i\}$
                \State \textbf{continue} \Comment{Mark failure, skip answering}
            \EndIf
            \State $a_i \gets M_{\text{rea}}(I_{\text{rea}}, z_i, \mathcal{E}_i)$
        \ElsIf{$\alpha_i = \texttt{answer}$}
            \State $a_i \gets M_{\text{rea}}(I_{\text{rea}}, z_i, \bigcup_{c_j \in \mathcal{S}} \mathcal{E}_j)$
        \EndIf
        \State $\mathcal{S} \gets \mathcal{S} \cup \{(c_i, a_i)\}$
        \State $\mathcal{P} \gets \mathcal{P} \setminus \{c_i\}$
    \EndFor
\EndWhile

\State \textbf{// Generate final answer}
\If{$\mathcal{P} = \emptyset$}
    \State $a_{\text{final}} \gets \text{Answer of last sub-question in } \mathcal{S}$
\Else
    \State $a_{\text{final}} \gets \text{FallbackReasoner}(q, P_q, \mathcal{S})$
\EndIf

\State \Return $a_{\text{final}}$
\end{algorithmic}
\end{algorithm}

%% file: sections/04_experiment.tex
\section{Experiments}
\input{tables/table_main}
\subsection{Experimental Setup}
\noindent\textbf{Datasets.}
We evaluate our method on three widely used multi-hop QA benchmarks: 2WikiMultihopQA~\cite{2wiki}, HotpotQA~\cite{yang2018hotpotqa}, and MuSiQue~\cite{trivedi2022musique}.
Following the data partitioning protocol of \citet{trivedi2023interleaving}, we extend the original test sets to 1,000 questions per dataset by randomly sampling an additional 500 examples from the training sets.
For the retrieval corpus, we construct a unified document collection by aggregating all gold supporting documents and distractor documents from each dataset  respectively, consistent with \citet{trivedi2023interleaving}. Furthermore, we conduct additional comparative experiments on a larger and noisier corpus in Section~\ref{exp:larger_corpora}.

\noindent\textbf{Evaluation Metrics.}
We report four standard metrics: Exact Match (EM), F1 score, Precision, and Recall.
For each question, EM is 1 if the prediction exactly matches the ground-truth answer and 0 otherwise; F1, Precision, and Recall are computed based on token-level overlap between the prediction and answer.
All metrics are then averaged over the dataset.
Following previous works~\cite{yang2018hotpotqa, dualrag}, all methods apply the same post-processing (e.g., string normalization and span extraction) before metric computation to ensure fairness.

\noindent\textbf{Baselines.}
We compare against a comprehensive set of baselines, categorized into three groups:

\begin{itemize}[leftmargin=*]
    \item \textbf{Non-retrieval}: We prompt LLMs with Chain-of-Thought (CoT)~\cite{wei2022chain} to perform multi-hop reasoning without external retrieval, assessing their intrinsic reasoning capability.
    
    \item \textbf{Single-step RAG}: We augment the RAG~\cite{lewis2020retrieval} pipeline with CoT prompting, where the model generates an answer based on documents retrieved in a single step using the original question. 
    
    \item \textbf{Iterative RAG}: We include state-of-the-art iterative frameworks, including SelfAsk~\cite{press2023measuring}, Iter-RetGen~\cite{shao2023enhancing}, GenGround~\cite{genground}, and DualRAG~\cite{dualrag}, which perform multi-turn retrieval and reasoning via sub-question decomposition or iterative refinement. These are discussed in Section \ref{related_work}.
\end{itemize}
To avoid confounding effects from prompt quality, we maintain consistent prompt styles across all baselines and STRIDE.

\subsection{Implementation Details}
To ensure a fair comparison, we cap the maximum number of iterations at 5 for all iterative methods.

\noindent\textbf{LLM Backbones.}
We primarily use \texttt{Qwen3-8B}~\cite{yang2025qwen3} as the backbone LLM and additionally evaluate \texttt{GPT-4o-mini} to assess the impact of model scale and architecture.
For all LLM calls, we fix the temperature to 1.0 and set the maximum generation length to 512.

\noindent\textbf{Retriever.}
We build dense vector indexes using FAISS-GPU~\cite{faiss} with the \texttt{Contriever}~\cite{contriever} encoder.
All baselines retrieve the Top-$k=5$ documents per query by default.
Notably, we set our method with $k=3$ on 2WikiMultihopQA and MuSiQue due to the strong performance.
We further analyze the sensitivity of performance to $k \in \{2,3,5,8\}$ in Section~\ref{app:topk}.

\noindent\textbf{Modular Fine-Tuning.}
We randomly sample 5,000 questions from each trainging set to form the tuning set.
Following the procedure in Section~\ref{sec:modular_tuning}, we collect filtered trajectories for each module, resulting in 371 samples for the Meta Planner, 217 for the Supervisor, 603 for the Extractor, and 1,119 for the Reasoner.
To reduce computational cost, we apply LoRA~\cite{lora} to fine-tune the four modules separately, with 2 epochs, lora rank 8, lora alpha 32, batch size 16, and learning rates of $3 \times 10^{-5}$ for DPO and $1 \times 10^{-4}$ for others.
All training runs are conducted on a  NVIDIA A6000 GPU with 2 hours in total.

\subsection{Main Results}

Table~\ref{tab:main_results} presents the main results across three MHQA benchmarks.
The results confirm the importance of external knowledge and iterative refinement in multi-hop reasoning, as retrieval-based methods consistently outperform non-retrieval baselines, and iterative RAG approaches generally achieve higher scores than single-step RAG.

Under the same training-free setting, STRIDE achieves the superior overall performance across both LLM backbones and most evaluation metrics.
Notably, on the highly compositional MuSiQue, STRIDE substantially outperforms all baselines: with GPT-4o-mini, it improves EM by 0.053 (0.33 vs. 0.277) and F1 by 0.068 (0.467 vs. 0.399) over DualRAG; similar gains are observed with Qwen3-8B (+0.059 EM, +0.072 F1).
This demonstrates that STRIDE’s explicit separation of strategy, control, and execution is particularly effective for complex, multi-step reasoning.

An interesting observation is the performance gap between LLM backbones across datasets.
On 2WikiMultihopQA, which features relatively structured reasoning paths, STRIDE with Qwen3-8B achieves higher EM (0.692) than with GPT-4o-mini (0.661), suggesting that Qwen3-8B’s strong instruction-following capability and output controllability lead to more precise answer generation.
In contrast, on the more challenging MuSiQue, GPT-4o-mini significantly outperforms Qwen3-8B (0.33 vs. 0.288 EM), indicating that its superior language understanding and generalization are critical when reasoning depth increases.
This highlights the complementary strengths of different LLMs and the adaptability of STRIDE across model families.

Furthermore, the fine-tuned variant (i.e., FT) provides consistent and substantial improvements.
For instance, on 2WikiMultihopQA with Qwen3-8B, FT boosts EM from 0.692 to 0.738 (+0.046), and on MuSiQue, it increases F1 from 0.401 to 0.448 (+0.047).
Notably, FT with Qwen3-8B closes the performance gap with GPT-4o-mini and even surpasses it on two datasets, validating that our modular fine-tuning strategy effectively leverages self-generated trajectories to enhance reasoning quality without external supervision.

Together, these results demonstrate that STRIDE enables robust and scalable multi-hop reasoning in complex scenarios.

\subsection{Ablation Study}
\input{tables/table_ablation}
We conduct comprehensive ablation studies to evaluate the contribution of each component in STRIDE and the effectiveness of modular fine-tuning. Results are reported in Tables~\ref{tab:ablation_core} and~\ref{tab:ablation_ft}.

\noindent\textbf{Ablation on the Framework.}
We first examine the impact of removing key modules from the inference pipeline.
\textbf{w/o Meta-Planner} disables the abstract reasoning strategy by removing all \textit{General Strategy}-related prompts, forcing the planner to directly output a concrete plan without high-level intent modeling.
This leads to significant performance drops across all datasets (e.g., $-$0.197 EM on 2WikiMultihopQA with Qwen3-8B), demonstrating that abstract planning provides crucial scaffolding for robust multi-hop decomposition.
\textbf{w/o Supervisor} replaces our process-aware supervisor with a conventional sequential execution pipeline, where sub-questions are answered in fixed order and each retrieval depends solely on the previous answer.
The resulting performance degradation, particularly on MuSiQue (-0.105 EM), highlights the importance of dynamic scheduling and dependency-aware control in handling complex reasoning graphs with parallel or conditional branches.
\textbf{w/o Extractor} skips atomic fact extraction and lets the reasoner directly generate answers from retrieved documents.
The consistent decline of performance suggests that explicit grounding of intermediate facts improves answer coherence and reduces hallucination.
Finally, \textbf{w/o Fallback} removes our recovery mechanism and falls back to standard RAG when execution fails.
The slight but consistent drops across datasets confirm that the fallback module enhances robustness, especially in complex scene where retrieval or reasoning may occasionally derail.
\input{tables/table_ablation_ft}

\noindent\textbf{Ablation on Modular Fine-Tuning.}
We further assess the contribution of fine-tuning each module by ablating one module’s training at a time (denoted as w/o MP-FT, w/o S-FT, etc.).
All variants underperform the full FT model, confirming that each module benefits from task-specific adaptation.
Notably, removing fine-tuning of the Reasoner (w/o R-FT) causes the largest drop (e.g., $-$0.037 EM on 2WikiMultihopQA), underscoring the importance of aligning the final answer generator with the structured reasoning process.
In contrast, ablating Meta-Planner fine-tuning (w/o MP-FT) has the mildest impact, suggesting that the abstract strategy is already well-captured by prompting, but still benefits from refinement.
These results validate our modular fine-tuning design: each component learns complementary skills, and joint adaptation yields the strongest performance.

\section{Comparative Analysis}
We conduct a systematic evaluation of STRIDE across eight complementary dimensions: from the effectiveness of \textit{meta-level planning} and the impact of \textit{rewrite actions}, to robustness under \textit{large retrieval corpora} and \textit{multi-hop reasoning depth}; from \textit{computational efficiency} and \textit{error attribution}, to \textit{iteration efficiency} and sensitivity to \textit{retrieval top-$k$}. Across all axes, STRIDE demonstrates consistent and significant advantages over existing iterative RAG methods.

\begin{figure}[htbp]
\centering
\includegraphics[width=\linewidth]{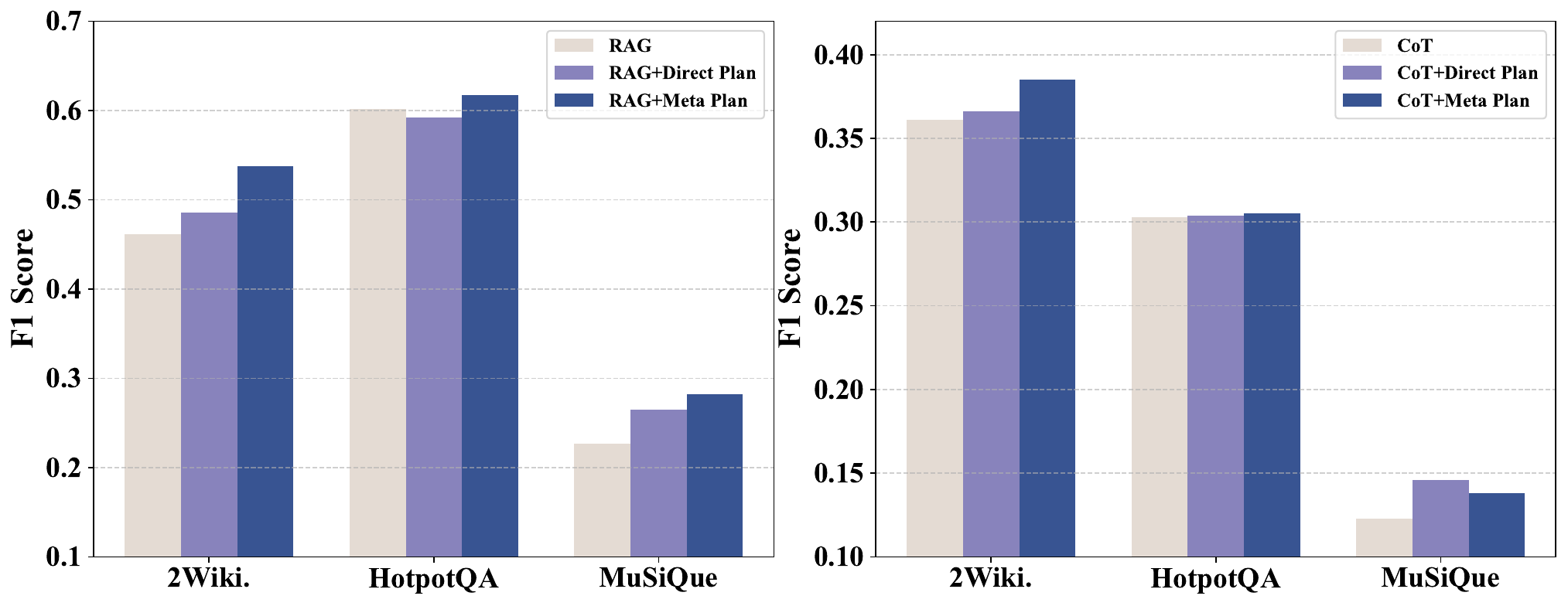}
\vspace{-20pt}
\caption{
F1 scores of RAG and CoT with different planning strategies. \textit{Meta Plan} generally outperforms \textit{Direct Plan}, with larger gains in the RAG setting, highlighting the value of abstract reasoning guidance for multi-hop QA.
}
\label{fig:plan_comparison}
\end{figure}

\subsection{Effectiveness of Meta-Level Planning}
To evaluate the benefit of abstract reasoning, we compare two planning paradigms in both retrieval-augmented (RAG) and retrieval-free (CoT) settings: (1) \textit{Direct Plan}, which generates concrete sub-questions only (equivalent to our w/o Meta-Planner ablation), and (2) \textit{Meta Plan}, which is produced by Meta-Planner.

As shown in Figure~\ref{fig:plan_comparison}, \textit{Meta Plan} generally improves F1 performance over \textit{Direct Plan} across all datasets. The gains are most pronounced in the RAG setting: on 2WikiMultihopQA, RAG with \textit{Meta Plan} achieves a +5 F1 improvement over \textit{Direct Plan}, demonstrating that high-level planning enables more effective decomposition and thus better retrieval and reasoning. Even on the more challenging MuSiQue, \textit{Meta Plan} maintains a consistent edge, confirming its robustness across task difficulties.
In contrast, under pure CoT (without retrieval), both planning strategies yield only marginal improvements over vanilla CoT, and the advantage of \textit{Meta Plan} is less pronounced. This suggests that structured, abstract planning primarily enhances performance when coupled with external evidence retrieval—where accurate question decomposition directly impacts the relevance and utility of retrieved passages.

Together, these results validate our core design principle: decoupling strategic intent from operational execution leads to more reliable, generalizable, and retrieval-aware reasoning plans.

\input{tables/table_rewrite}
\subsection{Impact of Rewrite Effectiveness}
The Supervisor module may trigger a \textit{rewrite} action to recover from flawed reasoning plans or insufficient evidence.
To evaluate the practical utility of this mechanism, we compute the fraction of rewrite-triggered questions that ultimately yield a non-zero F1 score, indicating that the intervention led to a meaningful answer.
As shown in Table~\ref{tab:rewrite_gain_ratio}, even without fine-tuning, STRIDE achieves a high rewrite success rate, confirming that the process-aware control logic effectively identifies recoverable failures.
More importantly, modular fine-tuning further enhances this capability: under STRIDE-FT, the success rate increases consistently.

\begin{figure}[htbp]
\centering
\includegraphics[width=\linewidth]{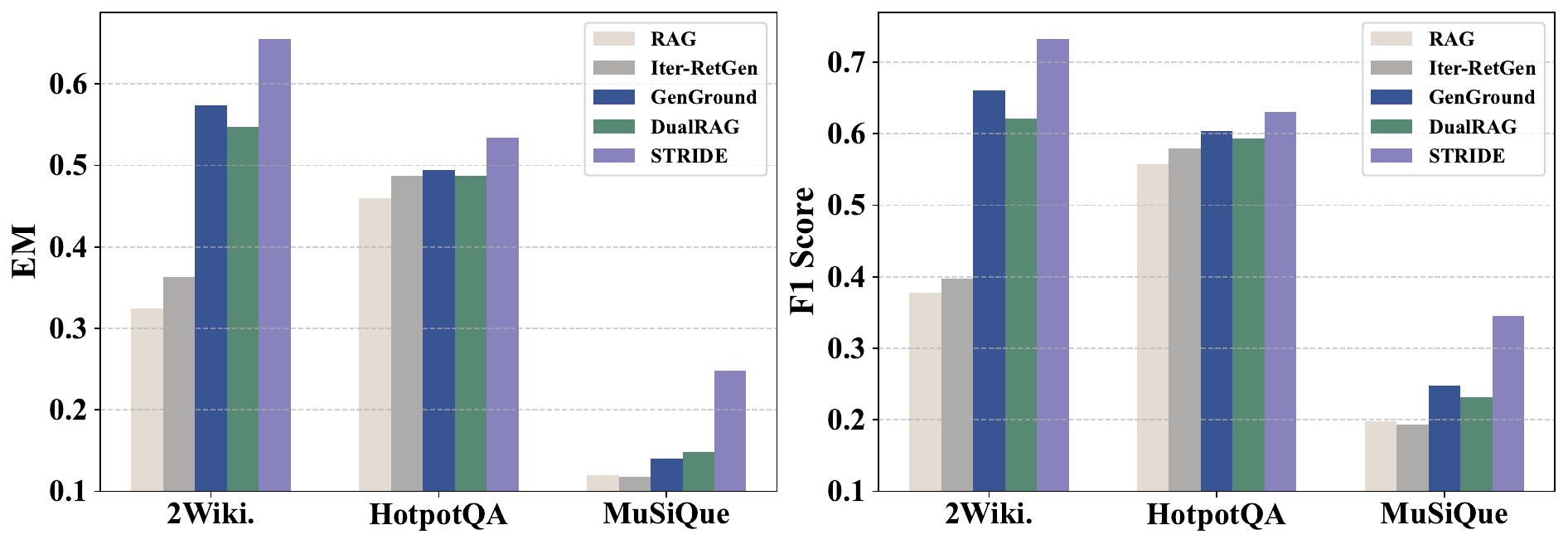}
\vspace{-20pt}
\caption{
Performance under expanded retrieval corpora (50k documents per dataset). 
STRIDE maintains the strongest performance and exhibits the smallest degradation, highlighting its robustness to retrieval noise.
}
\label{fig:5w_docs}
\vspace{-5pt}
\end{figure}

\subsection{Robustness under Larger Retrieval Corpora}
\label{exp:larger_corpora}
We expand the retrieval corpus to 50,000 documents per dataset to simulate realistic open-domain conditions with high noise and sparse evidence.
As shown in Figure~\ref{fig:5w_docs}, all methods suffer performance degradation due to the sparser signal-to-noise ratio.
Notably, \textbf{STRIDE achieves the best results across all datasets} in this challenging regime, with particularly large gains on the most complex MuSiQue.
Moreover, STRIDE exhibits the smallest relative drop compared to its performance on the original corpora, demonstrating exceptional robustness.
This resilience stems from STRIDE’s meta-level planning, which generates focused sub-questions that guide retrieval toward semantically relevant passages, even when the corpus is vast and noisy.
The results confirm that STRIDE’s structured reasoning not only improves accuracy but also enhances generalization to larger and noisier retrieval scenarios.

\begin{figure}[htbp]
\centering
\includegraphics[width=0.95\linewidth]{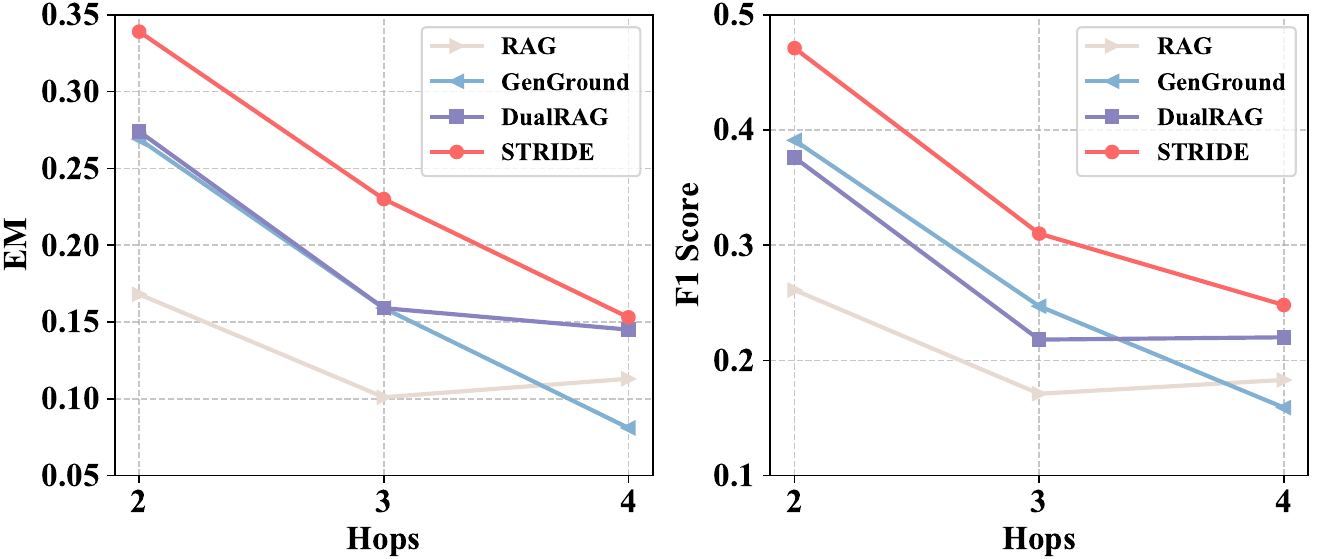}
\vspace{-10pt}
\caption{
Performance on MuSiQue with different hops. STRIDE shows consistent gains across all depths, highlighting its strength in complex reasoning.
}
\label{fig:hops}
\vspace{-10pt}
\end{figure}

\subsection{Performance across Reasoning Depths}
To investigate how reasoning complexity affects performance, we conduct experiments on MuSiQue grouped by the number of required inference hops.
As shown in Figure~\ref{fig:hops}, performance generally degrades as hop count increases, reflecting the growing challenge of maintaining reasoning fidelity over longer chains.
Notably, STRIDE consistently outperforms all baselines at every reasoning depth.
This trend indicates that STRIDE’s meta-level planning and dynamical scheduling mechanism are particularly effective at managing complex, multi-step reasoning, where error propagation and retrieval drift severely hinder conventional iterative RAG approaches.

\input{tables/table_efficiency}
\subsection{Efficiency Analysis}
Beyond accuracy, we evaluate the computational efficiency of STRIDE in terms of token consumption and inference time per question.
As shown in Table~\ref{tab:efficiency}, STRIDE achieves the highest performance while maintaining a favorable efficiency profile.
Compared to DualRAG, STRIDE reduces token usage by 54\%–71\% and inference time by 60\%–80\% across datasets, demonstrating that its hierarchical planning avoids the redundant retrieval-generation cycles typical of iterative RAG systems.
Although STRIDE incurs moderate overhead relative to the lightweight GenGround, it delivers a substantial gain in answer quality (+0.052 average F1).
This highlights STRIDE’s strong cost-accuracy trade-off: it provides state-of-the-art performance without prohibitive computational demands, making it well-suited for practical deployment.

\input{tables/table_error}
\subsection{Error Attribution Analysis}
We analyze failure modes on all incorrect predictions by first evaluating plan correctness using a strong LLM (DeepSeek-V3.2~\cite{liu2025deepseek}) under gold documents access. If the plan is flawed, we label it a \textit{Planning Error}. Otherwise, we check retrieval coverage: missing gold documents yield \textit{Retrieval Error}; otherwise, incorrect final answers are \textit{Execution Errors}.

As shown in Table~\ref{tab:error-attribution}, Meta-Plan reduces total failures across all datasets, with consistent decreases in retrieval and execution errors. Planning errors also decline on 2Wiki. and HotpotQA, indicating that strategic decomposition yields higher-quality plans for moderately complex questions.
On MuSiQue, planning errors remain nearly unchanged, yet total failures still decrease—suggesting that Meta-Plan produces more grounded and executable reasoning steps even when perfect planning remains challenging.

\begin{figure}[htbp]
\centering
\includegraphics[width=\linewidth]{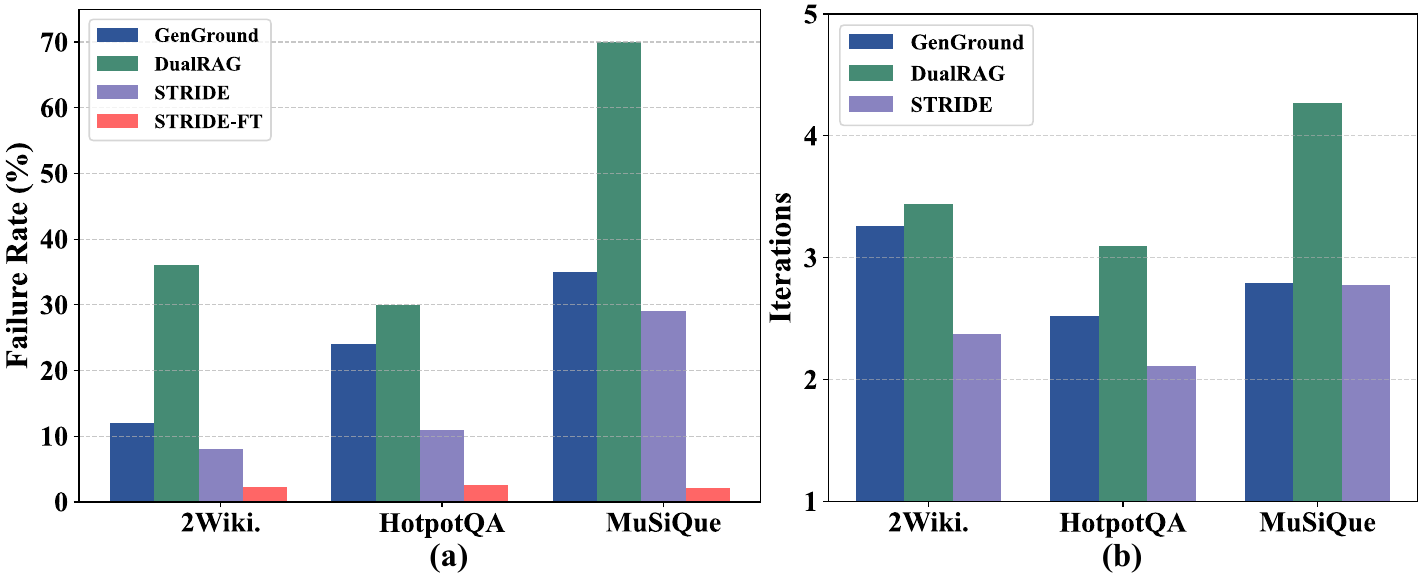}
\vspace{-20pt}
\caption{
(a) Failure rate (\%) and (b) average iteration count. 
Lower values indicate better robustness and higher reasoning efficiency, respectively. 
STRIDE achieves the best trade-off, and STRIDE-FT further reduces failure rates to near zero.
}
\label{fig:failure_and_iter}
\vspace{-5pt}
\end{figure}

\subsection{Execution Reliability and Iteration Efficiency}
We evaluate the reliability and efficiency of iterative RAG systems using two metrics: (a) \textit{failure rate}: the fraction of questions abandoned due to unrecoverable planning or retrieval errors, and (b) \textit{average iteration count}.
As shown in Figure~\ref{fig:failure_and_iter}, STRIDE achieves the lowest failure rate across all datasets, demonstrating that its meta-level planning and process-aware supervision effectively mitigate error propagation. 
Notably, with modular fine-tuning, STRIDE-FT further reduces the failure rate to below 3\% on all datasets, approaching near-perfect execution reliability.
Moreover, STRIDE requires fewer iterations on average than both baselines, indicating that dynamic scheduling avoids redundant refinement loops.

\begin{figure}[htbp]
\centering
\includegraphics[width=0.95\linewidth]{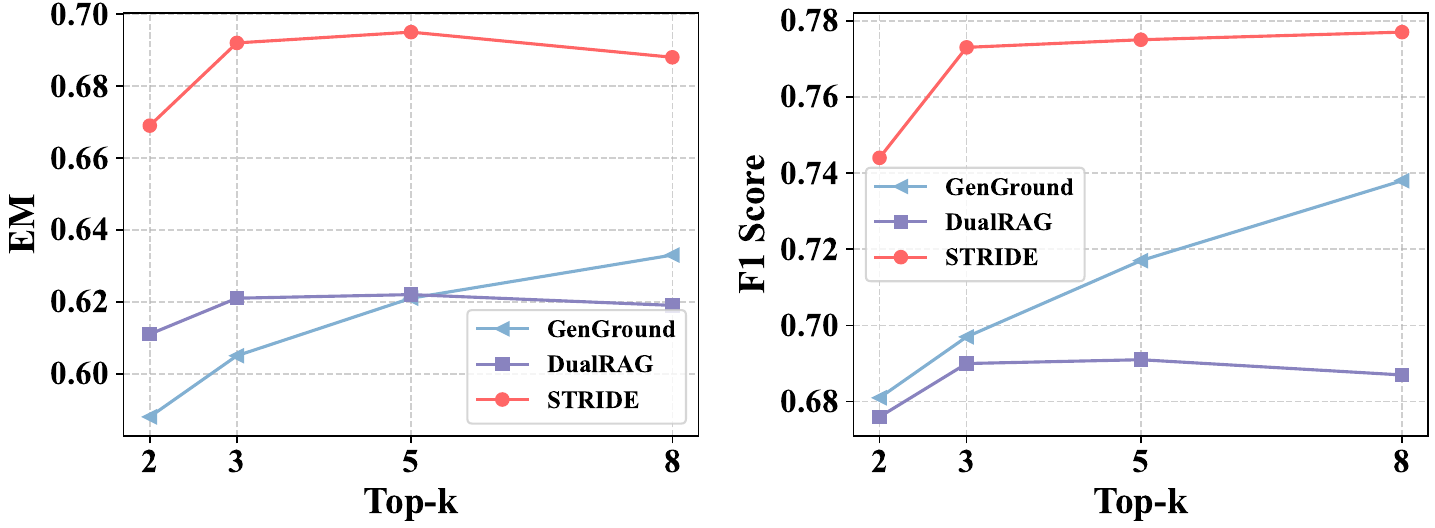}
\vspace{-10pt}
\caption{
Effect of retrieval top-$k$ on 2WikiMultihopQA. 
}
\label{fig:topk}
\vspace{-5pt}
\end{figure}

\subsection{Impact of Retrieval Top-$k$}
\label{app:topk}
Figure~\ref{fig:topk} shows the performance of all methods on 2WikiMultihopQA under varying top-$k$ values.
STRIDE achieves peak performance at $k=3$, with only marginal gains at $k=5$ or $8$.
In contrast, GenGround and DualRAG continue to improve up to $k=5$, justifying our choice to use $k=5$ for these baselines.
Notably, even with a smaller $k$, STRIDE outperforms both baselines at their best $k$ setting.
This demonstrates that STRIDE’s structured planning leads to more focused and effective retrieval, requiring fewer candidates to achieve superior accuracy.

%% file: tables/table_main.tex
\begin{table*}[htbp]
\centering
\caption{Evaluation results on multi-hop QA benchmarks. EM, F1, P, and R denote Exact Match, F1 score, Precision, and Recall respectively. Best and second-best results within each model group are highlighted in \textbf{bold} and \underline{underlined}, respectively.}
\label{tab:main_results}
\resizebox{0.95\textwidth}{!}{%
\begin{tabular}{lcccccccccccc}
\toprule
\multirow{2}{*}{\textbf{Method}} & \multicolumn{4}{c}{\textbf{2WikiMultihopQA}} & \multicolumn{4}{c}{\textbf{HotpotQA}} & \multicolumn{4}{c}{\textbf{MuSiQue}} \\
\cmidrule(lr){2-5} \cmidrule(lr){6-9} \cmidrule(lr){10-13}
 & EM & F1 & P & R & EM & F1 & P & R & EM & F1 & P & R \\
\midrule
& \multicolumn{12}{c}{\textit{GPT-4o-mini}} \\ \midrule
CoT           & 0.385 & 0.436 & 0.432 & 0.448 & 0.346 & 0.441 & 0.459 & 0.442 & 0.120 & 0.223 & 0.236 & 0.234 \\
RAG w/ CoT    & 0.484 & 0.552 & 0.547 & 0.573 & 0.522 & 0.633 & 0.646 & 0.669 & 0.186 & 0.308 & 0.313 & 0.335 \\
Iter-RetGen   & 0.446 & 0.517 & 0.505 & 0.557 & 0.551 & 0.668 & 0.676 & \textbf{0.723} & 0.156 & 0.289 & 0.267 & 0.444 \\
GenGround     & 0.467 & 0.628 & 0.594 & \underline{0.731} & 0.480 & 0.623 & 0.613 & \underline{0.705} & 0.230 & 0.372 & 0.355 & \underline{0.475} \\
DualRAG       & \underline{0.660} & \underline{0.720} & \underline{0.711} & 0.752 & \textbf{0.578} & \textbf{0.679} & \textbf{0.691} & 0.694 & \underline{0.277} & \underline{0.399} & \underline{0.393} & 0.481 \\
\rowcolor[gray]{0.92}\textbf{STRIDE}   & \textbf{0.661} & \textbf{0.777} & \textbf{0.744} & \textbf{0.837} & \underline{0.564} & \underline{0.671} & \underline{0.683} & 0.695 & \textbf{0.330} & \textbf{0.467} & \textbf{0.471} & \textbf{0.514} \\
\midrule
& \multicolumn{12}{c}{\textit{Qwen3-8B}} \\ \midrule
CoT           & 0.308 & 0.361 & 0.355 & 0.381 & 0.220 & 0.303 & 0.311 & 0.324 & 0.045 & 0.123 & 0.128 & 0.146 \\
RAG w/ CoT    & 0.397 & 0.460 & 0.451 & 0.503 & 0.501 & 0.602 & 0.616 & 0.625 & 0.144 & 0.227 & 0.234 & 0.262 \\
SelfAsk       & 0.299 & 0.362 & 0.351 & 0.394 & 0.449 & 0.520 & 0.529 & 0.542 & 0.166 & 0.240 & 0.239 & 0.269 \\
Iter-RetGen   & 0.413 & 0.455 & 0.453 & 0.467 & 0.519 & 0.619 & 0.633 & 0.634 & 0.148 & 0.239 & 0.251 & 0.264 \\
GenGround     & 0.621 & 0.717 & 0.702 & 0.760 & 0.523 & 0.628 & 0.636 & 0.664 & 0.218 & 0.325 & 0.331 & 0.353 \\
DualRAG       & 0.622 & 0.691 & 0.682 & 0.724 & \underline{0.552} & \underline{0.658} & \underline{0.666} & \textbf{0.701} & 0.229 & 0.316 & 0.321 & 0.349 \\
\rowcolor[gray]{0.92}\textbf{STRIDE} & \underline{0.692} & \underline{0.773} & \underline{0.757} & \textbf{0.815} & 0.549 & 0.653 & 0.662 & 0.681 & \underline{0.288} & \underline{0.401} & \underline{0.407} & \textbf{0.461} \\
\rowcolor[gray]{0.92}\textbf{STRIDE-FT} & \textbf{0.738} & \textbf{0.796} & \textbf{0.798} & \underline{0.806} & \textbf{0.601} & \textbf{0.691} & \textbf{0.723} & \underline{0.682} & \textbf{0.350} & \textbf{0.448} & \textbf{0.475} & \underline{0.448} \\
\bottomrule
\end{tabular}%
}
\end{table*}

%% file: tables/table_ablation.tex
\begin{table}[htbp]
\centering
\caption{Ablation study of STRIDE. Removing any module degrades performance, especially the Supervisor.}
\label{tab:ablation_core}
\resizebox{\linewidth}{!}{
\begin{tabular}{lcccccc}
\toprule
\multirow{2}{*}{\textbf{Variant}} & \multicolumn{2}{c}{\textbf{2Wiki.}} & \multicolumn{2}{c}{\textbf{HotpotQA}} & \multicolumn{2}{c}{\textbf{MuSiQue}} \\
\cmidrule(lr){2-3} \cmidrule(lr){4-5} \cmidrule(lr){6-7}
        & EM & F1 & EM & F1 & EM & F1 \\
\midrule
\rowcolor[gray]{0.92}STRIDE & 0.692 & 0.773 & 0.549 & 0.653 & 0.288 & 0.402 \\
w/o Meta-Planner       & 0.495 & 0.622 & 0.510 & 0.612 & 0.272 & 0.381 \\
w/o Supervisor      & 0.448 & 0.538 & 0.522 & 0.632 & 0.183 & 0.272 \\
w/o Extractor       & 0.673 & 0.764 & 0.531 & 0.645 & 0.271 & 0.371 \\
w/o Fallback        & 0.683 & 0.764 & 0.544 & 0.649 & 0.266 & 0.370 \\
\bottomrule
\end{tabular}}
\end{table}

%% file: tables/table_ablation_ft.tex
\begin{table}[htbp]
\centering
\caption{Ablation study on fine-tuning components of STRIDE. We remove each module (MP: Meta-Planner, S: Supervisor, E: Extractor, R: Reasoner) during fine-tuning.}
\label{tab:ablation_ft}
\resizebox{\linewidth}{!}{
\begin{tabular}{lcccccc}
\toprule
\multirow{2}{*}{\textbf{Variant}} & \multicolumn{2}{c}{\textbf{2Wiki.}} & \multicolumn{2}{c}{\textbf{HotpotQA}} & \multicolumn{2}{c}{\textbf{MuSiQue}} \\
\cmidrule(lr){2-3} \cmidrule(lr){4-5} \cmidrule(lr){6-7}
        & EM & F1 & EM & F1 & EM & F1 \\
\midrule
\rowcolor[gray]{0.92}STRIDE-FT  & 0.738 & 0.796 & 0.601 & 0.691 & 0.350 & 0.448 \\
w/o MP-FT       & 0.733 & 0.784 & 0.580 & 0.665 & 0.346 & 0.447 \\
w/o S-FT        & 0.734 & 0.783 & 0.588 & 0.679 & 0.322 & 0.426 \\
w/o E-FT        & 0.725 & 0.779 & 0.583 & 0.677 & 0.332 & 0.423 \\
w/o R-FT        & 0.701 & 0.774 & 0.573 & 0.671 & 0.312 & 0.415 \\
\bottomrule
\end{tabular}}
\end{table}

%% file: tables/table_rewrite.tex
\begin{table}[htbp]
\centering
\caption{Proportion of rewrite attempts that lead to performance improvement across diverse multi-hop QA datasets.}
\vspace{-5pt}
\label{tab:rewrite_gain_ratio}
\begin{tabular}{lcccc}
\toprule
Method & \textbf{2Wiki.} & \textbf{HotpotQA} & \textbf{MuSiQue} & \textbf{Average} \\
\midrule
STRIDE     & 50\% & 30\% & 41\% & 40.3\%\\
STRIDE-FT  & 65\% & 63\% & 48\% & 58.7\%\\
\bottomrule
\end{tabular}
\end{table}

%% file: tables/table_efficiency.tex
\begin{table}[t]
\centering
\caption{Token consumption and inference time (seconds) per question. ``Avg.'' is the abbreviation for ``average''.}
\vspace{-5pt}
\label{tab:efficiency}
\resizebox{\linewidth}{!}{
\begin{tabular}{lccccccc}
\toprule
\multirow{2}{*}{Method} &
\multicolumn{2}{c}{\textbf{2Wiki.}} &
\multicolumn{2}{c}{\textbf{HotpotQA}} &
\multicolumn{2}{c}{\textbf{MuSiQue}} &
\textbf{Avg.} \\
\cmidrule(lr){2-3} \cmidrule(lr){4-5} \cmidrule(lr){6-7} \cmidrule(lr){8-8}
& \#token$\downarrow$ & time$\downarrow$ & \#token$\downarrow$ & time$\downarrow$ & \#token$\downarrow$ & time$\downarrow$ & F1$\uparrow$\\
\midrule
GenGround & 5171 & 3.27 & 3452 & 2.64 & 4313 & 3.14 & 0.557 \\
DualRAG   & 16047 & 10.83 & 15148 & 11.05 & 25923 & 20.45 & 0.555 \\
\rowcolor[gray]{0.92} STRIDE    & 7436 & 4.37 & 7345 & 4.42 & 7656 & 4.11 & 0.609\\
\bottomrule
\end{tabular}}
\vspace{-5pt}
\end{table}

%% file: tables/table_error.tex
\begin{table}[t]
\centering
\caption{Error attribution on failed predictions across datasets. Counts reflect the number of incorrect samples attributed to each error type. Lower errors indicate higher performance.}
\vspace{-5pt}
\label{tab:error-attribution}
\resizebox{\linewidth}{!}{
\begin{tabular}{llcccc}
\toprule
Dataset & Method & Plan & Retrieval & Execution & Total  \\
\midrule
\multirow{2}{*}{2Wiki.} & Direct-Plan & 59 & 94 & 194 & 347 \\
 & Meta-Plan & 43 & 57 & 126 & 226 \\
\midrule
\multirow{2}{*}{HotpotQA} & Direct-Plan & 98 & 143 & 218 & 459 \\
 & Meta-Plan & 90 & 127 & 197 & 414  \\
\midrule
\multirow{2}{*}{MuSiQue} & Direct-Plan & 230 & 302 & 130 & 662 \\
 & Meta-Plan & 234 & 290 & 118 & 642  \\
\bottomrule
\end{tabular}
}
\vspace{-5pt}
\end{table}

%% file: sections/05_conclusion.tex
\section{Conclusion}
In this paper, we propose STRIDE, a decision-inspired framework for multi-hop question answering that separates reasoning into strategy, control, and execution. STRIDE first constructs an entity-agnostic reasoning skeleton before grounding it into concrete sub-questions, reducing sensitivity to lexical noise. Its Supervisor dynamically schedules sub-questions, not only supporting sequential, parallel, and fork-join patterns, but also adaptively chooses between retrieval and inference. To close the performance gap between open-source and closed-source models, we further introduce STRIDE-FT, which leverages self-executed trajectories to fine-tune open-source LLMs without human labels or stronger teachers. Extensive experiments show that STRIDE achieves robust and accurate multi-hop reasoning, and STRIDE-FT effectively enhances model performance, demonstrating the value of structured decision-inspired reasoning.